\documentclass[10pt,journal,compsoc]{IEEEtran}

\usepackage{amssymb} 
\usepackage{graphicx} 
\usepackage{amsmath}
\usepackage{hyperref}
\usepackage{textcomp}
\usepackage{multirow}
\usepackage{mathtools}

\usepackage{dsfont}
\usepackage{color}

\begin{document}

%
\title{Differential Recurrent Neural Network and Its Application for Human Activity Recognition}
%
%
%


\author{Naifan Zhuang,
		Guo-Jun Qi,~\IEEEmembership{Member,~IEEE},
        The Duc Kieu,
        and~Kien A. Hua,~\IEEEmembership{Fellow,~IEEE}
\thanks{Naifan Zhuang, Guo-Jun Qi, and Kien A. Hua are with the Department
of Computer Science, University of Central Florida, Orlando, FL, 32816. E-mail: zhuangnaifan@knights.ucf.edu, guojun.qi@ucf.edu, kienhua@cs.ucf.edu.}
\thanks{The Duc Kieu is with the Department of Computing and Information Technology,
University of the West Indies, St. Augustine, Trinidad and Tobago. E-mail: ktduc0323@yahoo.com.au.}
}


%
%

\markboth{IEEE Transactions on Pattern Analysis and Machine Intelligence}%
{Shell \MakeLowercase{\textit{et al.}}: Bare Demo of IEEEtran.cls for IEEE Journals}

%



\IEEEtitleabstractindextext{
\begin{abstract}

The Long Short-Term Memory (LSTM) recurrent neural network is capable of processing complex sequential information since it utilizes special gating schemes for learning representations from long input sequences. It has the potential to model any sequential time-series data, where the current hidden state has to be considered in the context of the past hidden states. This property makes LSTM an ideal choice to learn the complex dynamics present in long sequences. Unfortunately, the conventional LSTMs do not consider the impact of spatio-temporal dynamics corresponding to the given salient motion patterns, when they gate the information that ought to be memorized through time. To address this problem, we propose a differential gating scheme for the LSTM neural network, which emphasizes on the change in information gain caused by the salient motions between the successive video frames. This change in information gain is quantified by Derivative of States (DoS), and thus the proposed LSTM model is termed as differential Recurrent Neural Network (dRNN). 
In addition, the original work used the hidden state at the last time-step to model the entire video sequence. Based on the energy profiling of DoS, we further propose to employ the State Energy Profile (SEP) to search for salient dRNN states and construct more informative representations. The effectiveness of the proposed model was demonstrated by automatically recognizing human actions from the real-world 2D and 3D single-person action datasets. Extensive experimental studies were further performed on human group and crowd activity datasets to show the generalization ability of dRNN.
We point out that LSTM is a special form of dRNN. As a result, we have introduced a new family of LSTMs.
Our study is one of the first works towards demonstrating the potential of learning complex time-series representations via high-order derivatives of states.

\end{abstract}

}

\maketitle

\IEEEdisplaynontitleabstractindextext

%
\IEEEpeerreviewmaketitle

\section{Introduction}

\begin{figure*}
	\centering
		\includegraphics[width=1.0\linewidth]{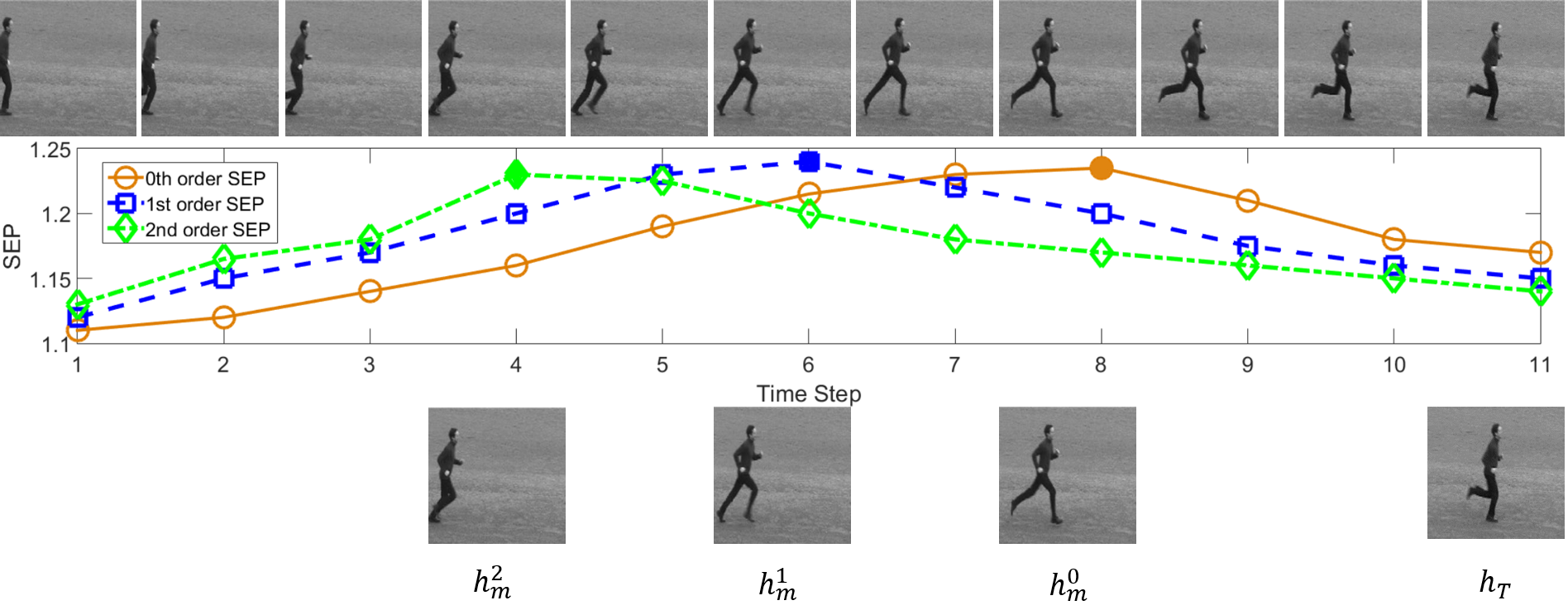}
	\caption{The top part shows a sequence of successive frames of "Running" behavior from the KTH dataset. The middle part plots State Energy Profile (SEP) for different orders of DoS. The solid dots denote the local maxima of SEP. The corresponding frames of local maxima are shown in the bottom part. Mean pooling is performed over hidden states at local maxima landmarks and the hidden state at the last time-step to generate a comprehensive sequence representation. }
	\label{fig:mmp}
\end{figure*}

\IEEEPARstart{R}{ecently}, Recurrent Neural Networks (RNNs) \cite{elman1990finding}, especially Long Short-Term Memory (LSTM) model \cite{hochreiter1997long}, have gained significant attention in solving many challenging problems involving time-series data, such as human activity recognition \cite{gregor2015draw, donahue2015long, grushin2013robust}, multilingual machine translation \cite{sutskever2014sequence, bahdanau2014neural}, multimodal translation \cite{venugopalan2014translating}, and robot control \cite{mayer2008system}. In these applications, learning an appropriate representation of sequences is an important step in achieving artificial intelligence.

Compared with existing spatio-temporal features \cite{klaser2008spatio, scovanner20073, chang2015heterogeneous, qi2012exploring} from the time-series data, RNNs use either a hidden layer \cite{schuster1997bidirectional} or a memory cell \cite{hochreiter1997long} to learn the time-evolving states which models the underlying dynamics of the input sequence. For example, \cite{donahue2015long} and \cite{baccouche2010action} have used LSTMs to model the video sequences to learn their long short-term dynamics. In contrast to the conventional RNN, the major component of LSTM is the memory cell which is modulated by three gates: input, forget, and output gates. These gates determine the amount of dynamic information entering or leaving the memory cell. The memory cell has a set of internal states, which store the information obtained over time. In this context, these internal states constitute a representation of an input sequence learned over time.

In many recent works, the LSTMs have shown tremendous potential in activity recognition tasks \cite{donahue2015long, grushin2013robust, baccouche2010action}. The existing LSTM models represent a video by integrating over time all the available information from each frame. However, we observed that for a human activity recognition task, not all frames contain salient spatio-temporal information which are discriminative to different classes of activities. Many frames contain non-salient motions which are irrelevant to the performed action.

This inspired us to develop a new family of LSTM model that automatically learns the dynamic saliency of the actions performed. The conventional LSTM fails to capture the salient dynamic patterns, since the gate units do not explicitly consider whether a frame contains salient motion information when they modulate the input and output of the memory cells. Thus, the model is insensitive to the dynamic evolution of the hidden states given the input video sequences. To address this problem, we propose the differential RNN (dRNN) model that learns these salient spatio-temporal representations of human activities. Specifically, dRNN models the dynamics of actions by computing different-orders of Derivative of State (DoS) that are sensitive to the spatio-temporal structure of input sequence. In other words, depending on the DoS, the gate units can learn the appropriate information that should be required to model the dynamic evolution of actions. 


In our prior work of dRNN \cite{veeriah2015differential}, we used hidden state at the last time-step to model the entire video sequence, which we call the Last-Hidden-State (LHS) method. In a very long sequence, the information learned from previous time steps decays gradually over time. Consequently, the hidden state at the last time-step tends be insufficient to model the whole sequence. In the meantime, we observed that each order of DoS energy could align a different level of motion saliency, as seen in Fig. \ref{fig:mmp}{\color{red}.}. We are motivated to address the above problem by digging into DoS to search for discriminative hidden states over different time-steps. 

Based on the observation of natural alignment of DoS energy and motion saliency, we now further propose to use State Energy Profile (SEP) to generate more discriminative and informative video representation. While DoS models the salient spatio-temporal representations of human activities, the motion energy intensity can be approximately estimated by the L2-norm of DoS. After plotting the energy curve of DoS over the time-steps, we can detect the local maxima landmarks of SEP. As shown in Fig. \ref{fig:mmp}{\color{red}.}, these landmarks indicate strong motion intensity at the corresponding time-steps and are more likely to provide discriminative information for recognizing human activities.
We then construct the video sequence representation based on the hidden states at those landmarks in addition to the hidden state at the last time-step.
To train the dRNN SEP model, we use truncated Back Propagation algorithm to prevent the exploding or diminishing errors through time \cite{hochreiter1997long}. In particular, we follow the rule that the errors propagated through the connections to those DoS nodes would be truncated once they leave the current memory cell.


To explore the potential of dRNN comprehensively, we have involved different orders of DoS to detect and capture the various levels of dynamical motion patterns for the dRNN model. The insight beneath is as follows. When modeling a moving object in a video, the $1^{st}$-order of DoS captures the velocity while the $2^{nd}$-order captures its acceleration. If we set the DoS to be $0^{th}$-order, which resembles the traditional LSTM, it should model the locality information. Until now, we can clearly see that the relationship between LSTM and dRNN. With a $0^{th}$-order of DoS, LSTM is a special case of dRNN. Higher-order dRNN captures not only locality information, but also velocity and acceleration information. Thus, we have introduced a new family of LSTMs; and LSTM is a special form of dRNN.

The effectiveness of dRNN is demonstrated on its application to human activity recognition. We show that dRNN can achieve the state-of-the-art performance on both 2D and 3D single-person action recognition datasets. Extensive experimental studies were further performed on human group and crowd activity datasets to show the generalization ability of dRNN. Specifically, dRNNs outperform the existing LSTM models on these activity recognition tasks, consistently achieving the better performance with the input sequences. Armed with SEP, dRNNs further enhance the experimental performances.
On the other hand, when compared with the other algorithms based on special assumptions about spatio-temporal structure of actions, the proposed general-purpose dRNN model can still reach competitive performance.

The remainder of this paper is organized as follows. In Section \ref{sec:related}{\color{red}.}, we review several related works on the human activity recognition problem. The background and details of RNNs and LSTMs are reviewed in Section \ref{sec:background}{\color{red}.}. We present the proposed dRNN model in Section \ref{sec:dRNN}{\color{red}.}. The experimental results are presented in Section \ref{sec:exp}{\color{red}.}. Finally, we conclude our work in Section \ref{sec:conclusion}{\color{red}.}.

\section{Related Work}
\label{sec:related}

Human activity recognition includes sub-problems of individual human action recognition and multi-person activity recognition. Multi-person activity recognition is further divided into group activity recognition and crowd analysis.

Action recognition has been a long-standing research problem in computer vision and pattern recognition community. This is a challenging problem due to the huge intra-class variance of actions performed by different actors at various speeds, in diverse environments (e.g., camera angles, lighting conditions, and cluttered background). To address this problem, many robust spatio-temporal representations have been constructed. For example, HOG3D \cite{klaser2010learning} uses the histogram of 3D gradient orientations to represent the motion structure over the frame sequences; 3D-SIFT \cite{scovanner20073} extends the popular SIFT descriptor to characterize the scale-invariant spatio-temporal structure for 3D video volume; actionlet ensemble \cite{wang2012mining} utilizes a robust approach to model the discriminative features from 3D positions of tracked joints captured by depth cameras. Although these descriptors have achieved remarkable success, they are usually engineered to model a specific spatio-temporal structure in an ad-hoc fashion. Recently, the huge success of deep networks in image classification \cite{krizhevsky2012imagenet} and speech recognition \cite{graves2014towards} has inspired many researchers to apply the deep neural networks, such as 3D Convolutional Neural Networks (3DCNNs) \cite{baccouche2011sequential} and Recurrent Neural Networks (RNNs) \cite{donahue2015long, baccouche2010action}, to action recognition. In particular, Baccouche \textit{et al.} \cite{baccouche2011sequential} developed a 3DCNN that extends the conventional CNN by taking space-time volume as input. On the contrary, \cite{donahue2015long} and \cite{baccouche2010action} used LSTMs to represent the video sequences directly, and modeled the dynamic evolution of the action states via a sequence of memory cells. Meanwhile, the existing approaches combine deep neural networks with spatio-temporal descriptors, achieving competitive performance. For example, in \cite{baccouche2011sequential}, an LSTM model takes a sequence of Harris3D and 3DCNN descriptors extracted from each frame as input, and the results on KTH dataset have shown the state-of-the-art performance \cite{baccouche2011sequential}.

Most existing approaches of group activity recognition are based on motion trajectories of group participants. Ni \emph{et~al.} \cite{ni2009recognizing} applied motion trajectory segments as inputs and used frequency responses of digital filters to represent the motion information. Zhu \emph{et~al.} \cite{zhu2011generative} considered motion trajectory as a dynamic system and used the Markov stationary distribution to acquire local appearance features as a descriptor of group action. 
Chu \emph{et~al.} \cite{chu2012new} designed an algorithm to model the trajectories as series of heat sources to create a heat map for representing group actions.
Cho \emph{et~al.} \cite{cho2015group} addressed the problem by using group interaction zones to detect meaningful groups to handle noisy information.
Cheng \emph{et~al.} \cite{cheng2014recognizing} proposed a layered model of human group action and represented activity patterns with both motion and appearance information. Their performance on NUS-HGA achieved an accuracy of 96.20\%. 
Zhuang \textit{et al.} \cite{zhuang2017group} used a combination of Deep VGG network \cite{simonyan2014very} and stacked LSTMs. Their model complexity is high and has a large chance of overfitting. In this case, their model is trained on augmented data thus cannot be fairly compared with other methods.

Recent methods for crowd scene understanding mostly analyze crowd activities based on motion features extracted from trajectories/tracklets of objects \cite{shao2014scene, hassner2012violent, su2016crowd, mousavi2015analyzing}. Marsden \textit{et al}. \cite{marsden2016holistic} studied scene-level holistic features using tracklets to solve real-time crowd behavior anomaly detection problem. \cite{marsden2016holistic} holds the state-of-the-art performance for the Violent-Flows dataset. Su \textit{et al}. \cite{su2016crowd} used tracklet-based features and explored Coherent LSTM to model the nonlinear characteristics and spatio-temporal motion patterns in crowd behaviors. The trajectory/tracklet feature contains more semantic information, but the accuracy of trajectories/tracklets dictates the performance of crowd scene analysis. In extremely crowded areas, tracking algorithms could fail and generate inaccurate trajectories. The general-purpose dRNN does not require such input, holding the potential for more sequence-related applications.

\section{Background}
\label{sec:background}

In this section, we review in detail the recurrent neural network as well as its variant, long short-term memory model. Readers who are familiar with them might skip to the next section directly.

\subsection{Recurrent Neural Networks}

Traditional Recurrent Neural Networks (RNNs) \cite{elman1990finding} model the dynamics of an input sequence of frames $\{\textbf{x}_t \in \mathds{R}^{m} | t = 1, 2, ..., T\}$ through a sequence of hidden states
$\{\textbf{h}_t \in \mathds{R}^{n} | t = 1, 2, ..., T\}$, thereby learning the spatio-temporal structure of the input sequence. For instance, a classical RNN model uses the following recurrent equation
\begin{equation}
	\textbf{h}_t = \tanh(\textbf{W}_{hh}\textbf{h}_{t-1} + \textbf{W}_{hx}\textbf{x}_{t} + \textbf{b}_h),
\end{equation}
to model the hidden state $\textbf{h}_t$ at time $t$ by combining the information from the current input $\textbf{x}_t$ and the past hidden state $\textbf{h}_{t-1}$. The hyperbolic tangent $\tanh(\cdot)$ in the above equation is an activation function with range [-1, 1]. $\textbf{W}_{hh}$ and $\textbf{W}_{hx}$ are two mapping matrices to the hidden states, and $\textbf{b}_h$ is the bias vector.

The hidden states will then be mapped to an output sequence $\{\textbf{y}_t \in \mathds{R}^{k} | t = 1, 2, ..., T\}$ as
\begin{equation}{\label{eq:out__}}
	\textbf{y}_t = \tanh(\textbf{W}_{yh}\textbf{h}_{t} + \textbf{b}_y),
\end{equation}
where each $\textbf{y}_t$ represents a 1-of-$k$ encoding of the confidence scores on $k$ classes of human activities. This output can then be transformed to a vector of probabilities $\textbf{p}_t$ by the softmax function as
\begin{equation}{\label{eq:softmax}}
	p_{t,c} = \frac{\exp(y_{t,c})}{\sum_{l=1}^k\exp(y_{t,l})},
\end{equation}
where each entry $p_{t,c}$ is the probability of frame $t$ belonging to class $c \in \{1,...,k\}$.

\subsection{Long Short-Term Memory}

Due to the exponential decay in retaining the context information from video frames, the aforementioned classical RNNs are limited in learning the long-term representation of sequences. To overcome this limitation, Long Short-Term Memory (LSTM) \cite{hochreiter1997long}, a variant of RNN, has been designed to exploit and find the long-range dependency between the input sequences and output labels.

Specifically, LSTM consists of a sequence of memory cells, each containing an internal state $\textbf{s}_{t}$ to store the memory of the input sequence up to time $t$. In order to retain the memory with respect to a context in a long sequence, three types of gate units are incorporated into LSTMs to regulate what information enters and leaves the memory cells over time. These gate units are activated by a nonlinear function of input/output sequences as well as internal states, which makes LSTM powerful enough to model dynamically changing contexts.

Formally, an LSTM cell has the following gates:

(i) Input gate $\textbf{i}_{t}$ regulates the degree to which the input information would enter the memory cell to affect the internal state $\textbf{s}_t$ at time $t$. The activation of the gate has the following recurrent form:
\[\textbf{i}_{t} = \sigma(\textbf{W}_{is}\textbf{s}_{t-1} + \textbf{W}_{ih}\textbf{h}_{t-1} + \textbf{W}_{ix}\textbf{x}_{t} + \textbf{b}_i),\]
where the sigmoid $\sigma(\cdot)$ is an activation function in the range [0,1], with 0 meaning the gate is closed and 1 meaning the gate is completely open; $\textbf{W}_{i*}$ are the mapping matrices and $\textbf{b}_i$ is the bias vector.

(ii) Forget gate $\textbf{f}_t$ modulates the previous state $\textbf{s}_{t-1}$ to control its contribution to the current state. It is defined as
\[ \textbf{f}_t = \sigma(\textbf{W}_{fs}\textbf{s}_{t-1} + \textbf{W}_{fh}\textbf{h}_{t-1} + \textbf{W}_{fx}\textbf{x}_{t} + \textbf{b}_f),\] with the mapping matrices $\textbf{W}_{f*}$ and the bias vector $\textbf{b}_f$.

With the input and forget gate units, the internal state $\textbf{s}_t$ of each memory cell can be updated as below:
\begin{equation}\label{eq:inter_state}
 \textbf{s}_t = \textbf{f}_t \otimes \textbf{s}_{t-1} + \textbf{i}_t \otimes \tanh(\textbf{W}_{sh}\textbf{h}_{t-1} + \textbf{W}_{sx}\textbf{x}_t + \textbf{b}_s),  
\end{equation}
where $\otimes$ stands for element-wise product. 

(iii) Output gate $\textbf{o}_t$ gates the information output from a memory cell which would influence the future states of LSTM cells. It is defined as
\[\textbf{o}_t = \sigma(\textbf{W}_{os}\textbf{s}_{t} + \textbf{W}_{oh}\textbf{h}_{t-1} + \textbf{W}_{ox}\textbf{x}_{t} + \textbf{b}_o).\]
Then the hidden state of a memory cell is output as 
\begin{equation}\label{eq:out}
\textbf{h}_t = \textbf{o}_t \otimes \tanh(\textbf{s}_t).
\end{equation}
In brief, LSTM proceeds by iteratively applying Eq. (\ref{eq:inter_state}) and Eq. (\ref{eq:out}) to update the internal state $\textbf{s}_t$ and the hidden state $\textbf{h}_t$. In the process, the input gate, forget gate, and output gate play an important role in controlling the information entering and leaving the memory cell. More details about LSTMs can be found in \cite{hochreiter1997long}.

\section{Derivative of States and State Energy Profile}
\label{sec:dRNN}

\subsection{Derivative of States}

\begin{figure}
	\centering
		\includegraphics[width=1.0\linewidth]{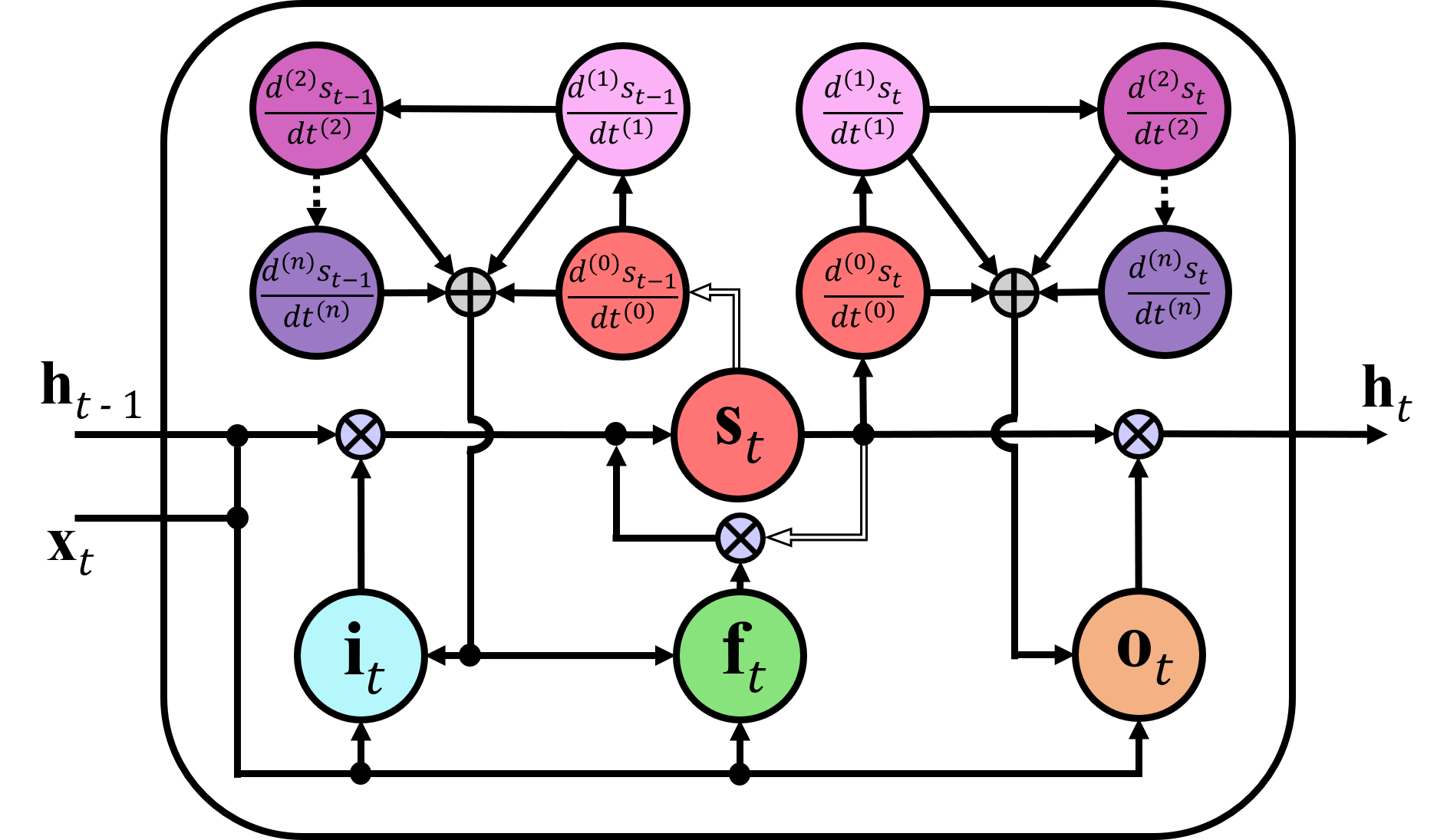}
	\caption{Architecture of dRNN model at time-step $t$. Best viewed in color.}
	\label{fig:dlstm_archi}
\end{figure}

For the task of activity recognition, not all video frames contain salient patterns to discriminate between different classes of activities. Many spatio-temporal descriptors, such as 3D-SIFT \cite{scovanner20073} and HoGHoF \cite{laptev2008learning}, have been proposed to localize and encode the salient spatio-temporal points. They detect and encode the spatio-temporal points related to salient motions of the objects in video frames, revealing the important dynamics of actions.

In this paper, we propose a novel LSTM model to automatically learn the dynamics of human activities, by detecting and integrating the salient spatio-temporal sequences. Although traditional LSTM neural network is capable of processing complex sequential information, it might fail to capture the salient motion patterns because the gate units do not \textit{explicitly} consider the impact of dynamic structures present in input sequences. This makes the conventional LSTM model inadequate to learn the evolution of human activity states.

As the internal state of each memory cell contains the accumulated information of the spatio-temporal structure, the Derivative of States (Dos) $\frac{d\textbf{s}_t}{dt}$ quantifies the change of information at each time $t$. In other words, a large magnitude of DoS indicates that a salient spatio-temporal structure containing the informative dynamics is caused by an abrupt change of human activity state. In this case, the gate units should allow more information to enter the memory cell to update its internal state. On the other hand, when the magnitude of DoS is small, the incoming information should be gated out of the memory cell so that the internal state would not be affected by the current input. Therefore, DoS can be used as a factor to gate the information flow into and out of the internal state of the memory cell over time.

Moreover, we can involve higher-orders of DoS to detect and capture the higher-order dynamic patterns for the dRNN model. For example, when modeling a moving object in a video, the $1^{st}$-order of DoS captures the velocity and the $2^{nd}$-order captures its acceleration. These different orders of DoS will enable dRNN to better represent the dynamic evolution of human activities in videos.

Fig. \ref{fig:dlstm_archi} illustrates the architecture of the proposed dRNN model. Formally, we have the following recurrent equations to control the gate units with the DoS up to order $N$:
\begin{equation}\label{eq:input}
	\textbf{i}_{t} = \sigma(\sum_{n=0}^{N}\textbf{W}^{(n)}_{id}\frac{d^{(n)}\textbf{s}_{t-1}}{dt^{(n)}} + \textbf{W}_{ih}\textbf{h}_{t-1} + \textbf{W}_{ix}\textbf{x}_{t} + \textbf{b}_i),
\end{equation}

\begin{equation}\label{eq:forget}
	\textbf{f}_{t} = \sigma(\sum_{n=0}^{N}\textbf{W}^{(n)}_{fd}\frac{d^{(n)}\textbf{s}_{t-1}}{dt^{(n)}} + \textbf{W}_{fh}\textbf{h}_{t-1} + \textbf{W}_{fx}\textbf{x}_{t} + \textbf{b}_f),
\end{equation}

\begin{equation}\label{eq:output}
	\textbf{o}_{t} = \sigma(\sum_{n=0}^{N}\textbf{W}^{(n)}_{od}\frac{d^{(n)}\textbf{s}_{t}}{dt^{(n)}} + \textbf{W}_{oh}\textbf{h}_{t-1} + \textbf{W}_{ox}\textbf{x}_{t} + \textbf{b}_o),
\end{equation}
where $\frac{d^{(n)}\textbf{s}_{t}}{dt^{(n)}}$ is the $n^{th}$-order DoS. Until now, it can be clearly seen that when $N = 0$, the dRNN model resembles the conventional LSTM. Therefore, LSTM is a special form of dRNN.

It is worth pointing out that we do not use the derivative of input as a measurement of salient dynamics to control the gate units. The derivative of input would amplify the unwanted noises which are often contained in the input sequence. In addition, this derivative of input only represents the local dynamic saliency, in contrast to the long short-term change in the information gained over time. For example, a similar movement which has occurred several frames ago and been stored by LSTM, could be treated as a novel salient motion using derivative of inputs. On the contrary, DoS does not have this problem because the internal state $\textbf{s}_t$ has long-term memory of the past motion pattern and would not treat the same motion as a salient pattern if it has previously occurred.


Since dRNN is defined in the discrete-time domain, the $1^{st}$-order derivative $\frac{d\textbf{s}_t}{dt}$, as the velocity of information change, can be discretized as the difference of states:
\begin{equation}{\label{eq:vel}}
	\textbf{v}_t \triangleq \frac{d\textbf{s}_t}{dt} \doteq \textbf{s}_t - \textbf{s}_{t-1},
\end{equation}
for simplicity \cite{epperson2013introduction}.

Similarly, we consider the $2^{nd}$-order of DoS as the acceleration of information change. It can be discretized as:
\begin{equation}{\label{eq:acc}}
	\textbf{a}_t \triangleq \frac{d^2\textbf{s}_t}{dt^2} \doteq \textbf{v}_t - \textbf{v}_{t-1} = \textbf{s}_t - 2\textbf{s}_{t-1} + \textbf{s}_{t-2}.
\end{equation}

In this paper, we only consider the first two orders of DoS. Higher orders can be derived in a similar way.

\subsection{State Energy Profile}

In our prior work, we used the hidden state at the last time-step to represent the entire sequence. 
Even though dRNN is good at modeling long input sequential data, the hidden state at the last time-step still might not be enough to summarize the complex dynamic evolution of human activities. While dRNN processes the input frames sequentially, hidden state information learned from previous frames decays gradually over a very long sequence. It tends to be suboptimal to only use the hidden state from last time-step for video representation.

As mentioned above, DoS represents information gain of internal states between consecutive video frames. The L2-norm of DoS can approximately estimate the motion energy intensity of the human activities.
We name the above estimated energy intensity over all time steps as State Energy Profile (SEP). Thus, SEP can be used to locate hidden states with salient information.
Specifically, a salient spatio-temporal frame could yield a large magnitude of DoS, which corresponds to a local maximum in SEP. Therefor, in terms of SEP, we can determine whether information gain is strong at certain time-steps and thus local maxima of SEP indicate the most informative hidden states. Moreover, we involve different orders of SEP to detect different levels of dynamic patterns.
Formally we compute the SEP, denoted as $\mathcal{E}_t^n$, in terms of $n^{th}$-order DoS as follows:
\begin{equation}{\label{eq:sep}}
	\mathcal{E}_t^n = \|\frac{d^{(n)}\textbf{s}_{t}}{dt^{(n)}}\|_2, (n = 0, 1, ..., N).
\end{equation}

In this paper, we consider up to $2^{nd}$-order of SEP. Again, since the dRNN model is defined in discrete-time domain, SEP $\mathcal{E}_t^n$ with different orders are discretized as below:
\begin{equation}{\label{eq:sep0}}
	\mathcal{E}_t^0 = \|\frac{d^{(0)}\textbf{s}_{t}}{dt^{(0)}}\|_2 = \|\textbf{s}_{t}\|_2,
\end{equation}
\begin{equation}{\label{eq:sep1}}
	\mathcal{E}_t^1 = \|\frac{d^{(1)}\textbf{s}_{t}}{dt^{(1)}}\|_2 = \|\textbf{v}_{t}\|_2,
\end{equation}
\begin{equation}{\label{eq:sep2}}
	\mathcal{E}_t^2 = \|\frac{d^{(2)}\textbf{s}_{t}}{dt^{(2)}}\|_2 = \|\textbf{a}_{t}\|_2.
\end{equation}

In order to locate maxima landmarks, we plot SEP curves on different orders. Note that since LSTMs have long-term memory and controllable virtues, SEP curves are smoothed without the need of filtering. This would not be observed if derivative of input is applied instead of DoS. Based on the SEP curves, we then detect local maxima landmarks of SEP for different orders. Again, those local maxima correspond to time steps of salient and strong motion patterns. To be more specific, local maxima landmarks of $\mathcal{E}_t^1$ correspond to the time-steps with high velocity and maxima landmarks of $\mathcal{E}_t^2$ correspond to the time-steps with high acceleration. 
As seen in Fig. \ref{fig:mmp}{\color{red}.}, the action of Running exhibits the most informative pattern when the person in the video is of high speed and/or high acceleration. These moments correspond to the local maxima landmarks of SEP.

Based on the above observation, we construct the set of hidden states corresponding to the local maxima of SEP as $\{\textbf{h}^n_{m_1}, \textbf{h}^n_{m_2}, ... , \textbf{h}^n_{m_u}\}, n = 0, 1, ..., N$. Here $N$ is the highest order of DoS, and $u$ is the number of local maxima landmarks for order $n$. Since LSTM cell aggregates over all the time-steps, the hidden state at the last time-step contains the overall information of the video sequence.
We then form the Candidate Set of hidden states by adding the hidden state at the last time-step to the above set. Candidate Set is then denoted as $\{\textbf{h}^n_{m_1}, \textbf{h}^n_{m_2}, ... , \textbf{h}^n_{m_u}, \textbf{h}_T\}$, here $T$ denotes the last time-step.
To suppress the unwanted noise, mean pooling is then performed over the hidden states in the above Candidate Set to create the final representation for the entire sequence:
\begin{equation}{\label{eq:hidden_mean}}
	\textbf{h}_\tau = \mu(\{\textbf{h}^n_{m_1}, \textbf{h}^n_{m_2}, ... , \textbf{h}^n_{m_u}, \textbf{h}_T\}),
\end{equation}
where $\mu$ denotes mean pooling$, n = 0, 1, ..., N$.


To support our motivation of learning LSTM representations based on the SEP method, we illustrate example frames of Running activity from KTH dataset with SEP signal and local maxima landmarks in Fig. \ref{fig:mmp}{\color{red}.}. It shows that local maxima of SEP correspond to most intense motion frames, where the most salient and informative hidden states are located.
In this specific case, the maxima of SEP signal correspond to the moments when the person in the video runs at high speed or acceleration.
Interestingly, some human activities such as Walking and Running, exhibit regular motion periodicity \cite{wang2014person}. SEP also increases the possibility of finding aligned video frames.

To better understand the potential of SEP pooling strategy, we discuss the relationship between SEP and other pooling methods. There are currently two frequently used pooling methods. Mean pooling, by averaging all time-step hidden states, statistically summarizes the information collected from all previous frames and thus has a proper representation of the states. Due to the smoothing nature of mean pooling, the salient human activity information acquired by the neural network could be lost in the process. Max pooling is better at selecting salient signals and often generates more discriminative representation, but it is subject to noise. The proposed SEP pooling strategy, tailored for the dRNN model, combines the advantages of mean pooling and max pooling while minimizing the drawbacks.

\subsection{Learning Algorithm}

With the above recurrent equations, for a human activity video sequence including $T$ time-step features, the SEP dRNN model proceeds in the following order at time step $t$:
\begin{itemize}
	\item Compute input gate activation $\textbf{i}_t$ and forget gate activation $\textbf{f}_t$ by Eq. (\ref{eq:input}) and Eq. (\ref{eq:forget}).
	\item Update state $\textbf{s}_t$ with $\textbf{i}_t$ and $\textbf{f}_t$ by Eq. (\ref{eq:inter_state}).
    \item Compute discretized DoS $\{\frac{d^{(n)}\textbf{s}_{t-1}}{dt^{(n)}}|n=0,1,...,N\}$ up to $N^{th}$-order at time $t$, e.g. Eq. (\ref{eq:vel}) and Eq. (\ref{eq:acc}).
	\item Compute output gate $\textbf{o}_t$ by Eq. (\ref{eq:output}).
	\item Output $\textbf{h}_t$ gated by $\textbf{o}_t$ from memory cell by Eq. (\ref{eq:out}).
    \item (For frame level prediction) Output the label probability $\textbf{p}_t$ by applying Eq. (\ref{eq:out__}) and softmax Eq. (\ref{eq:softmax}) to $\textbf{h}_t$.
	\item (For sequence level prediction) Compute discretized SEP $\{\mathcal{E}_t^n|n=0,1,...,N\}$ using Eq. (\ref{eq:sep0}), Eq. (\ref{eq:sep1}), and Eq. (\ref{eq:sep2}).
\end{itemize}

To learn the model parameters of dRNN, we define a loss function to measure the deviation between the target class $c_t$ and $\textbf{p}_t$ at time $t$:
\[\ell(\textbf{p}_t,c_t) = -\log p_{t,{c_t}}.\]
If an individual level $c_t$ is given to each frame $t$ in the sequence, we can minimize the cumulative loss over the sequence:
\[\sum_{t=1}^{T}\ell(\textbf{p}_t,c_t).\]

For an activity recognition task, the label of activity is often given at the video level, so we mainly consider sequence level prediction. After completing the above recurrent steps for a video sequence, video representation using SEP $\textbf{h}_{\tau}$ is then computed by Eq. (\ref{eq:hidden_mean}). The sequence level class probability $\textbf{p}$ is generated by computing the output of dRNN with Eq. (\ref{eq:out__}) and applying the softmax function with Eq. (\ref{eq:softmax}). For a given training label $c$, the dRNNs can be trained by minimizing the loss function below, i.e.
\[\ell(\textbf{p},c) = -\log p_{c}.\]

The loss function can be minimized by Back Propagation Through Time (BPTT) \cite{hochreiter1997long}, which unfolds an LSTM model over several time-steps and then runs the back propagation algorithm to train the model. To prevent back-propagated errors from decaying or exploding exponentially, we use the truncated BPTT according to \cite{hochreiter1997long} to learn the model parameters. Specifically, in our model, errors are not allowed to re-enter the memory cell once they leave it through the DoS nodes.

  Formally, we assume the following truncated derivatives of gate activations:
\[ \frac{\partial \textbf{i}_t}{\partial \textbf{v}_{t-1}} \circeq 0, \frac{\partial \textbf{f}_t}{\partial \textbf{v}_{t-1}} \circeq 0, \frac{\partial \textbf{o}_t}{\partial \textbf{v}_{t}} \circeq 0,\]
and
\[ \frac{\partial \textbf{i}_t}{\partial \textbf{a}_{t-1}} \circeq 0, \frac{\partial \textbf{f}_t}{\partial \textbf{a}_{t-1}} \circeq 0, \frac{\partial \textbf{o}_t}{\partial \textbf{a}_{t}} \circeq 0,\]
where $\circeq$ stands for the truncated derivatives. The details about the implementation of the truncated BPTT can be found in \cite{hochreiter1997long}.

\section{Experiments and Results}
\label{sec:exp}

We compare the performance of the proposed method with the state-of-the-art LSTM and non-LSTM methods present in existing literature on human activity datasets.

\subsection{Datasets}
The proposed method is evaluated on individual human action recognition datasets: KTH and MSR Action3D datasets, as well as group and crowd activity recognition datasets: NUS-HGA and Violent-Flow datasets.

\textbf{KTH dataset.} We choose KTH dataset \cite{schuldt2004recognizing} because it is a \textit{de facto} benchmark for evaluating action recognition algorithms. This makes it possible to directly compare with the other algorithms. There are two KTH datasets: KTH-1 and KTH-2, which both consist of six action classes: walking, jogging, running, boxing, hand-waving, and hand-clapping. The actions are performed several times by 25 subjects in four different scenarios: indoors, outdoors, outdoors with scale variation, and outdoors with different clothes. The sequences are captured over homogeneous background with a static camera recording 25 frames per second. Each video has a resolution of 160 $\times$ 120, and lasts for about 4 seconds on KTH-1 dataset and one second for KTH-2 dataset. There are 599 videos in the KTH-1 dataset and 2,391 video sequences in the KTH-2 dataset.

\textbf{MSR Action3D dataset.} The MSR Action3D dataset \cite{li2010action} consists of 567 depth map sequences performed by 10 subjects using a depth sensor similar to the Kinect device. The resolution of each video is 320 $\times$ 240 and there are 20 action classes where each subject performs each action two or three times. The actions are chosen in the context of gaming. They cover a variety of movements related to arms, legs, torso, etc. This dataset has a lot of noise in the joint locations of the skeleton as well as high intra-class variations and inter-class similarities, making it a challenging dataset for evaluation among the existing 3D dataset. We follow a similar experiment setting from \cite{wang2012mining}, where half of the subjects are used for training and the other half are used for testing. This setting is much more challenging than the subset one used in \cite{li2010action} because all actions are evaluated together and the chance of confusion is much higher.

\textbf{NUS-HGA dataset.} We choose the NUS-HGA dataset \cite{ni2009recognizing} as it is a well-collected benchmark dataset for evaluating group activity recognition techniques. The NUS-HGA dataset includes 476 video clips covering six group activity classes: Fight, Gather, Ignore, RunInGroup, StandTalk, and WalkInGroup. Each instance involves 4-8 persons. The sequences are captured over different backgrounds with a static camera recording of 25 frames per second. Each video clip has a resolution of 720 $\times$ 576 and lasts around 10 seconds. 
We follow the experiment setting from \cite{cheng2014recognizing} and evaluate our method via five-fold cross validation.

\textbf{Violent-Flows dataset.} The Violent-Flows (VF) dataset \cite{hassner2012violent} is a real-world video footage of crowd violence, along with standard benchmark protocols designed for violent/non-violent classification. The Violent-Flows dataset includes 246 real-world videos downloaded from YouTube. The shortest clip duration is 1.04 seconds, the longest clip is 6.52 seconds, and the average length of the video is 3.6 seconds. We follow the standard 5-fold cross-validation protocol in \cite{hassner2012violent}.


\subsection{Feature Extraction}

We are using densely sampled HOG3D features to represent each frame of video sequences from the KTH dataset. Specifically, we uniformly divide the 3D video volumes into a dense grid, and extract the descriptors from each cell of the grid. The parameters for HOG3D are the same as the one used in \cite{klaser2008spatio}. 
The size of the descriptor is 1,000 per cell of grid, and there are 58 such cells in each frame, yielding a 58,000-dimensional feature vector per frame. We apply PCA to reduce the dimension to 2000, retaining 90\% of energy among the principal components, to construct a compact input into the dRNN model.

For 3D action dataset, MSR Action3D, a depth sensor like Kinect provides an estimate of 3D joint coordinates of body skeleton, and the following features were extracted to represent MSR Action3D depth sequences - (1) Position: 3D coordinates of the 20 joints obtained from the skeleton map. These 3D coordinates were then concatenated resulting in a 60-dimensional feature per frame; (2) Angle: normalized pair-wise angles. The normalized pair-wise angles were obtained from 18 joints of the skeleton map. The two feet joints were not included. This resulted in a 136-dimensional feature vector per frame; (3) Offset: offset of the 3D joint positions between the current and the previous frame \cite{zhu2013fusing}. These offset features were also computed using the 18 joints from the skeleton map resulting in a 54-dimensional feature per frame; (4) Velocity: histogram of the velocity components obtained from point cloud. This feature was computed using the 18 joints as in the previous cases resulting in a 162-dimensional feature per frame; (5) Pairwise joint distances: The 3D coordinates obtained from the skeleton map were used to compute pairwise joint distances with the center of the skeleton map resulting in a 60-dimensional feature vector per frame. For the following experiments, these five different features were concatenated to result in a 583-dimensional feature vector per frame.


Same as the KTH dataset, we choose HOG3D feature for NUS-HGA and Violent-Flows datasets and similar procedures of feature extraction are performed on these two datasets. After PCA is applied, NUS-HGA and Violent-Flows have a feature dimension of 300 and 500, respectively.

\subsection{Architecture and Training}
The architecture of the dRNN models trained on the above datasets is shown in Table \ref{tab:archit}{\color{red}.}. For the sake of fair comparison, we adopt the same architecture for both orders of dRNN models. We can see that the number of memory cell units is smaller than the input units on all datasets. This can be interpreted as follows. The sequence of a human activity video often forms a continuous pattern embedded in a low-dimensional manifold of the input space. Thus, a lower-dimension state space is sufficient to capture the dynamics of such patterns. The number of output units corresponds to the number of classes in the datasets.

\begin{table}
	\centering
		\begin{tabular}{ c | c | c | c | c }
		\hline
			 & \textbf{KTH} & \textbf{MSR} & \textbf{NUS} & \textbf{VF} \\ 
		\hline
			Input Units & 2000 & 583 & 300 & 500 \\
			State Units & 1500 & 400 & 200 & 400 \\
			Output Units & 6 & 20 & 6 & 2 \\
		\hline	
		\end{tabular}
	\caption{Architectures of the proposed model used on the datasets.}
	\label{tab:archit}
\end{table}

During training, the learning rate of the BPTT algorithm is set to 0.0001. The objective loss continuously decreases over 50 epochs. Usually after 40 epochs, the dRNN model begins to converge.


\subsection{Results on KTH Dataset}

There are several different evaluation protocols used on the KTH dataset in literature. This can result in as large as 9\% difference in performance across different experiment protocols as reported in \cite{gao2010comparing}. For fair comparison, we follow the cross-validation protocol \cite{baccouche2011sequential}, in which we randomly select 16 subjects to train the model, and test over the remaining 9 subjects. The performance is reported by the average across five such trials.

\begin{table}
	\centering
		\begin{tabular}{ c| c | c | c }
		\hline
			\textbf{Dataset} & \textbf{LSTM Model} & \textbf{LHS} & \ \textbf{SEP} \\ 
		\hline
        	\multirow{4}{4em}{KTH-1} & conventional LSTM + HOF \cite{grushin2013robust} & 87.78 & -- \\
                                     & conventional LSTM + HOG3D & 89.93 & -- \\
                                     & $1^{st}$-order dRNN + HOG3D  & 93.28 & 93.94 \\
                                     & $2^{nd}$-order dRNN + HOG3D & 93.96 & 94.78 \\
        \hline    
            \multirow{4}{4em}{KTH-2} & conventional LSTM + HOF \cite{baccouche2011sequential} & 87.78 & -- \\
                                     & conventional LSTM + HOG3D & 87.32 & -- \\
                                     & $1^{st}$-order dRNN + HOG3D  & 91.98 & 92.73 \\
                                     & $2^{nd}$-order dRNN + HOG3D & 92.12 & 92.93 \\
		\hline	
		\end{tabular}
	\caption{Performance comparison of LSTM models on the KTH-1 and KTH-2 datasets.}
	\label{tab:dRNN_kth}
\end{table}

\begin{figure}
	\centering
		\includegraphics[width=1.0\linewidth]{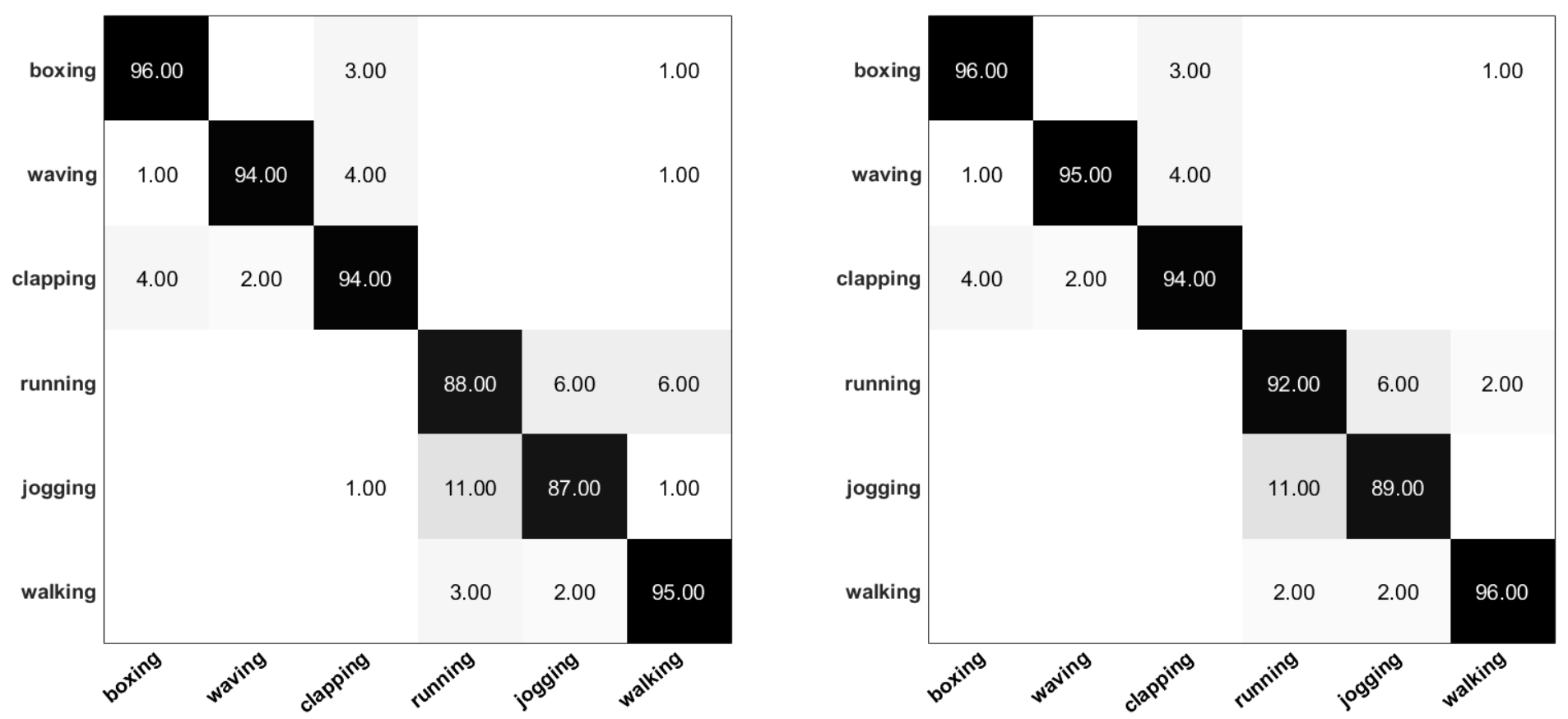}
	\caption{Confusion Matrices on the KTH-1 dataset obtained by the $2^{nd}$-order dRNN model using LHS (left) and SEP (right).}
	\label{fig:kth_cf}
\end{figure}

First, we compare the dRNN model with the conventional LSTM model in Table \ref{tab:dRNN_kth}{\color{red}.}. Here we report the cross-validation accuracy on both KTH-1 and KTH-2 datasets. In addition, Fig. \ref{fig:kth_cf} shows the confusion matrix obtained by the $2^{nd}$-order dRNN model on the KTH-1 dataset. This confusion matrix is computed by averaging over five trials in the above cross-validation protocol. The performance of the conventional LSTM has been reported in literature \cite{grushin2013robust, baccouche2011sequential}. We note that these reported accuracies often vary with different types of features. Thus, a fair comparison between different models can only be made with the same type of input feature.

For the dRNN model, we report the accuracy with up to the $2^{nd}$-order of DoS. Table \ref{tab:dRNN_kth} shows that with the same HOG3D feature, the proposed dRNN models outperform the conventional LSTM model, which demonstrate the effectiveness of dRNN. Utilizing DoS, dRNN explicitly models the change in information gain caused by the salient motions between the successive frames, thus can benefit the recognition process. For both the first and second orders of dRNNs, SEP consistently achieves better performance for KTH-1 and KTH-2 datasets. Based on DoS, SEP selected the most discriminative hidden states over all time steps, thus can generate more comprehensive representation than the hidden state at the last time-step. In the meanwhile, we can see that the $2^{nd}$-order dRNN yields a better accuracy than its $1^{st}$-order counterpart. Although higher order of DoS might improve the accuracy further, we do not report the result since it becomes trivial to simply add more orders of DoS into dRNN, and the improved performance might not even compensate for the increased computational cost. Moreover, with an increased order of DoS, more model parameters would have to be learned with the limited training examples. This tends to cause overfitting problem, making the performance stop improving or even begin to degenerate after the order of DoS reaches a certain number. Therefore, for most of practical applications, the first two orders of dRNN should be sufficient. 


Baccouche \textit{et al}. \cite{baccouche2011sequential} reported an accuracy of 94.39\% and 92.17\% on KTH-1 and KTH-2 datasets, respectively. It is worth noting that they used a combination of 3DCNN and LSTM, where 3DCNN plays a crucial role in reaching such performance. Actually, 3DCNN model alone can reach an accuracy of 91.04\% and 89.40\% on KTH-1 and KTH-2 datasets reported in \cite{baccouche2011sequential}. On the contrary, they reported that the LSTM with Harris3D feature only achieved 87.78\% on KTH-2, as compared with 92.93\% accuracy obtained by $2^{nd}$-order dRNN with SEP using HOG3D feature. In Table \ref{tab:dRNN_kth}{\color{red}.}, under a fair comparison with the same feature, the dRNN models of both orders outperform their LSTM counterpart with the same HOG3D feature.



\begin{figure*}
	\centering
		\includegraphics[width=1.0\linewidth]{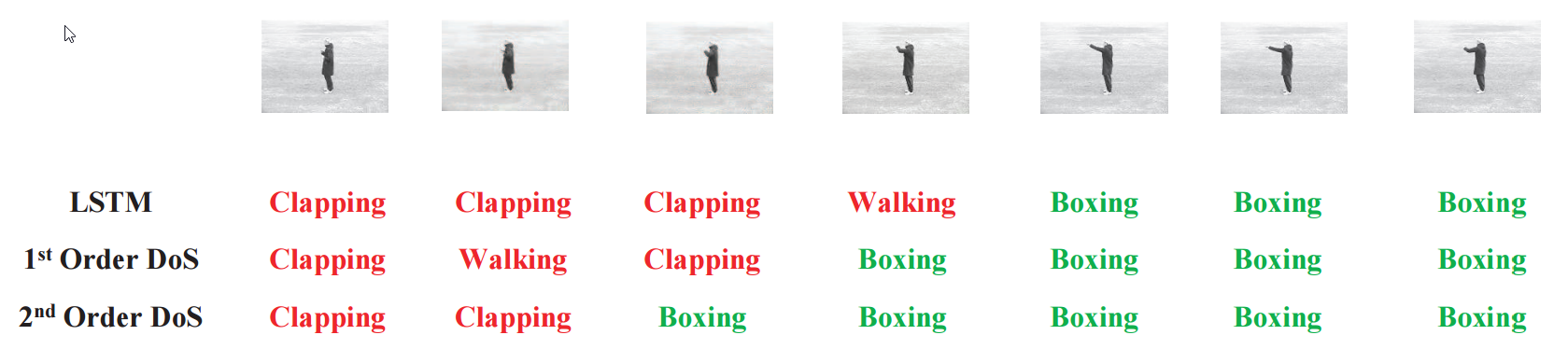}
	\caption{Frame-by-frame prediction of action category over time with LSTM and SEP dRNNs. Best viewed in color.}
	\label{fig:fbf}
\end{figure*}

To support our motivation of learning LSTM representations based on the dynamic change of states evolving over frames, we illustrate the predictions over time in Fig. \ref{fig:fbf}{\color{red}.}. From the result, we found that as time evolves, the proposed dRNNs are faster in learning the salient dynamics for predicting the correct action category than the LSTMs. Moreover, the $2^{nd}$-order DoS is better than $1^{st}$-order of DoS in learning the salient features.

\begin{table}
	\centering
		\begin{tabular}{ c | c | c }
		\hline
			\textbf{Dataset} & \textbf{Method} & \textbf{Accuracy (\%)} \\ 
		\hline
        	\multirow{4}{4em}{KTH-1} & Rodrigues \textit{et al}. \cite{rodriguez2008action} & 81.50 \\ 
									 & Jhuang \textit{et al}. \cite{jhuang2007biologically} & 91.70 \\ 
                                     & Schidler \textit{et al}. \cite{schindler2008action} & 92.70 \\
									 & 3DCNN \cite{baccouche2011sequential} & 91.04 \\
                                     & 3DCNN + LSTM \cite{baccouche2011sequential} & 94.39 \\
        \hline
        	\multirow{4}{4em}{KTH-2} & Ji \textit{et al}. \cite{ji20133d} & 90.20 \\ 
									 & Taylor \textit{et al}. \cite{taylor2010convolutional} & 90.00 \\ 
                                     & Laptev \textit{et al}. \cite{laptev2008learning} & 91.80 \\
									 & Dollar \textit{et al}. \cite{dollar2005behavior} & 81.20 \\
                                     & 3DCNN \cite{baccouche2011sequential} & 89.40 \\
                                     & 3DCNN + LSTM \cite{baccouche2011sequential} & 92.17 \\
		\hline	
		\end{tabular}
	\caption{Cross-validation accuracy over five trials obtained by the other compared algorithms on KTH-1 and KTH-2 datasets}
	\label{tab:KTH}
\end{table}

We also show the performance of the other non-LSTM state-of-the-art approaches in Table \ref{tab:KTH}{\color{red}.}. Many of these compared algorithms focus on the action recognition problem, relying on the special assumptions about the spatio-temporal structure of actions. They might not be applicable to model the other type of sequences which do not satisfy these assumptions. In contrast, the proposed dRNN model is a general-purpose model, not being tailored to specific type of action sequences. This also makes it competent on 3D action recognition and even more challenging tasks such as group and crowd activity recognition, which we will show below.

\subsection{Results on MSR Action3D Dataset}

\begin{table}
	\centering
		\begin{tabular}{ c | c | c }
		\hline
			\textbf{LSTM Model} & \textbf{LHS} & \ \textbf{SEP} \\ 
		\hline
        	conventional LSTM & 87.78 & 89.12 \\
            $1^{st}$-order dRNN & 91.40 & 92.74 \\
            $2^{nd}$-order dRNN & 92.03 & 92.91 \\
		\hline	
		\end{tabular}
	\caption{Performance comparison of LSTM models on the MSR Action3D dataset.}
	\label{tab:dRNN_msr}
\end{table}

\begin{table}
	\centering
		\begin{tabular}{ c | c  }
		\hline
			\textbf{Method} & \textbf{Accuracy (\%)} \\ 
		\hline
		    Actionlet Ensemble \cite{wang2012mining} & 88.20 \\
			HON3D \cite{oreifej2013hon4d} & 88.89 \\
			DCSF \cite{xia2013spatio} & 89.30\\
			Lie Group \cite{vemulapalli2014human} & 89.48\\
		\hline	
		\end{tabular}
	\caption{Performances of the other compared methods on MSR Action3D dataset.}
	\label{tab:MSR}
\end{table}

\begin{figure}
	\centering
		\includegraphics[width=1.1\linewidth]{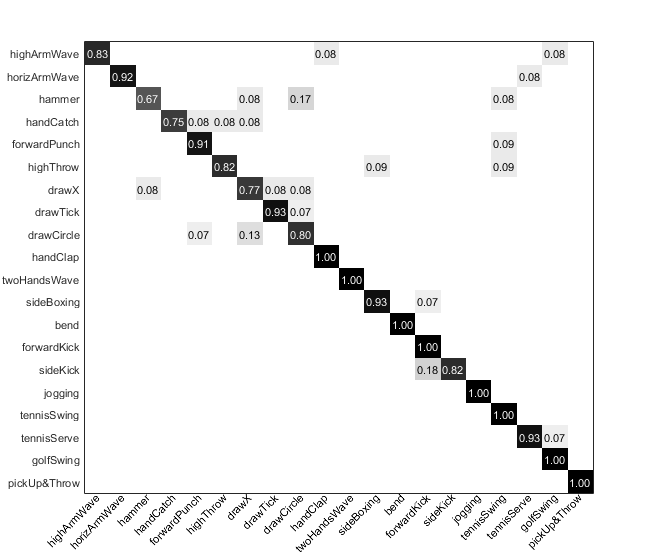}
	\caption{Confusion Matrix on the MSR Action3D dataset obtained by the $2^{nd}$-order dRNN model using SEP.}
	\label{fig:msr_cf}
\end{figure}

Table \ref{tab:dRNN_msr} and Table \ref{tab:MSR} compare the results on MSR Action3D dataset. Fig. \ref{fig:msr_cf} shows the confusion matrix by the $2^{nd}$-order dRNN model with SEP. The results are obtained by following exactly the same experimental setting in \cite{wang2012mining}, in which half of actor subjects are used for training and the rest are used for testing. This is in contrast to another evaluation protocol in literature \cite{li2010action} which splits across 20 action classes into three subsets and performs the evaluation within each individual subset. This evaluation protocol is more challenging because it is evaluated over all 20 action classes with no common subjects in training and testing sets.

From the results, all the dRNN models outperform the conventional LSTM algorithm with the same feature. Also dRNN models using SEP outperform dRNNs using LHS. By taking into consideration of not only the hidden state at the last time-step, which might have lost certain information from early frames due to exponential decay, dRNNs with SEP can generate more informative representation.

In the meanwhile, dRNN models perform competitively as compared with the other algorithms. We notice that the Super Normal Vector (SNV) model \cite{yang2014super} has reported an accuracy of 93.09\% on MSR Action3D dataset. However, this model is based on a special assumption about the 3D geometric structure of the surfaces of depth image sequences. Thus, this approach is a very special model for solving 3D action recognition problem. This is contrary to dRNN as a general model without any specific assumptions on the dynamic structure of the video sequences.

In brief, through the experiments on both 2D and 3D human action datasets, we show the competitive performance of dRNN compared with both LSTM and non-LSTM models. This demonstrates its wide applicability in representing and modeling the dynamics of both 2D and 3D action sequences, irrespective of any assumptions on the structure of video sequences. In the following sections, we demonstrate the application of dRNN on even more challenging tasks of multi-person activity recognition.

\subsection{Results on NUS-HGA Dataset}

For the sake of fair comparison, we follow \cite{cheng2014recognizing} and evaluate our method via a five-fold cross validation. We compare our method with conventional LSTM and previous group activity recognition methods in Table \ref{tab:dRNN_nus} and Table \ref{tab:HGA}{\color{red}.}, respectively. 

\begin{table}
	\centering
		\begin{tabular}{ c | c | c }
		\hline
			\textbf{LSTM Model} & \textbf{LHS} & \ \textbf{SEP} \\ 
		\hline
        	conventional LSTM & 93.48 & 94.71 \\
            $1^{st}$-order dRNN & 96.36 & 97.43 \\
            $2^{nd}$-order dRNN & 97.37 & 98.95 \\
		\hline	
		\end{tabular}
	\caption{Performance comparison of LSTM models on the NUS-HGA dataset.}
	\label{tab:dRNN_nus}
\end{table}

\begin{table}
	\centering
		\begin{tabular}{ c | c  }
		\hline
			\textbf{Method} & \textbf{Accuracy (\%)} \\ 
		\hline
		  Ni \emph{et~al.} \cite{ni2009recognizing} & 73.50 \\
			Zhu \emph{et~al.} \cite{zhu2011generative} & 87.00 \\
			Cho \emph{et~al.} \cite{cho2015group} & 96.03\\
			Cheng \emph{et~al.} \cite{cheng2014recognizing} (MF) & 93.20\\
			Cheng \emph{et~al.} \cite{cheng2014recognizing} (MAF) & 96.20\\
		\hline	
		\end{tabular}
	\caption{Performances of the other compared methods on NUS-HGA dataset. MF indicates motion feature fusion and MAF indicates motion and appearance feature fusion.}
	\label{tab:HGA}
\end{table}

As shown in Table \ref{tab:dRNN_nus}{\color{red}.}, all the dRNN models outperforms the conventional LSTM model. This again shows the effectiveness of explicitly analyzing dynamic structures between successive frames using DoS. Comparing between the dRNN models, $2^{nd}$-order dRNN achieves better performance than $1^{st}$-order dRNN. As mentioned above, $2^{nd}$-order dRNN analyzes not only the velocity, but also the acceleration information between internal states, which enlarge its ability for understanding more complex spatio-temporal dynamics. For the same order of dRNN, SEP outperforms LHS more than 1\%, which can be viewed as large margin considering that the performance is already very high. This again demonstrates the effectiveness of SEP.

In addition, dRNN models generally achieve better performance than the other non-LSTM methods. It is worth nothing that $2^{nd}$-order dRNN with SEP achieves almost 99\% recognition accuracy, outperforming \cite{cheng2014recognizing} by 2.75\%. Traditional solutions for group activity algorithms need human supervision to acquire accurate human object trajectories from the videos. According to \cite{cheng2014recognizing}, they acquired human bounding boxes using existing tracking tools, which requires manual annotation for bounding box initialization. This constraint prevents their method from being used in automatic or real-time applications. On the other hand, dRNN models can outperform these traditional methods without human manual annotation, enabling a much broader applications for group behavior analysis.


\subsection{Results on Violent-Flows Dataset}




To evaluate our method on the Violent-Flows dataset, we follow the standard 5-fold cross-validation protocol in \cite{hassner2012violent} and report the results in terms of mean accuracy in Table \ref{tab:dRNN_vf} and Table \ref{tab:VF}{\color{red}.}. 

From Table \ref{tab:dRNN_vf}{\color{red}.}, we can see that both the first and second orders of dRNN models outperform the traditional LSTM. This indicates that for even more complex scenario such as crowd behavior, DoS can still model the dynamical structures present in the video sequences more effectively. $2^{nd}$-order of dRNN consistently achieves slightly better performance than $1^{st}$-order of dRNN. As mentioned before, high orders of dRNNs have more model parameters to be trained. For a relatively small dataset such as Violent-Flows, overfitting problem tends to be more severe. 
On the other hand, Table \ref{tab:dRNN_vf} shows that the proposed SEP pooling strategy further boosts the performances, which again demonstrates the effectiveness of State Energy Profile over the Last Hidden State strategy.
In addition, dRNN models outperform the other non-LSTM state-of-the-art methods too, as seen in Table \ref{tab:VF}{\color{red}.}.

\begin{table}
	\centering
		\begin{tabular}{ c | c | c }
		\hline
			\textbf{LSTM Model} & \textbf{LHS} & \ \textbf{SEP} \\ 
		\hline
        	conventional LSTM & 83.78 & 84.92 \\
            $1^{st}$-order dRNN & 86.37 & 87.83 \\
            $2^{nd}$-order dRNN & 86.98 & 87.84 \\
		\hline	
		\end{tabular}
	\caption{Performance comparison of LSTM models on the Violent-Flows dataset.}
	\label{tab:dRNN_vf}
\end{table}

\begin{table}
	\centering
		\begin{tabular}{ c | c  }
		\hline
			\textbf{Method} & \textbf{Accuracy (\%)} \\ 
        \hline 
		    Violent Flows \cite{hassner2012violent} & 81.30 \\
			Common Measure \cite{mousavi2015crowd} & 81.50 \\
			Hist of Tracklet \cite{mousavi2015analyzing} & 82.30 \\
			Substantial Derivative \cite{mohammadi2015violence} & 85.43 \\
            Holistic Features \cite{marsden2016holistic} & 85.53 \\
		\hline	
		\end{tabular}
	\caption{Performances of the other compared methods on Violent-Flows dataset.}
	\label{tab:VF}
\end{table}

\subsection{SEP vs. other pooling techniques}

To show the advantage of SEP pooling method, we compare the results of SEP against other pooling strategies, e.g. Last-Hidden-State (LHS), Mean-Pooling, Max-Pooling, on KTH-1 and NUS-HGA datasets in Table \ref{tab:Pooling}{\color{red}.}. For fair comparison, all the results are acquired using $2^{nd}$-order dRNN model. LHS, used in our prior work, simply employs the hidden state at the last time-step to optimize the parameters of dRNN neural networks. Mean pooling/max pooling indicates that mean pooling/max pooling is performed over all the hidden states to generate representation of the input video. These three methods serve as baselines to compare with our proposed SEP pooling method.

The LHS method achieves the lowest performance compared to other methods. Since the information learned from previous time steps decays slowly over a very long sequence, the last time-step hidden state is unable to summarize the most distinguishing motion patterns from successive states. Mean pooling of all hidden states achieves a slightly better performance than using only the last hidden states, as it statistically averages and summarizes the information collected from all states. In the process of smoothing, mean pooling might lose the salient information acquired by the neural network. Max pooling, on the contrary, is good at selecting discriminative and salient information from the human behaviors. Because of this, max pooling over hidden states gets better results than LHS and mean pooling methods. On the other hand, max pooling is subject to motion noise. If a noisy and strong motion happens during the video sequence, max pooling could misclassify the human activity.

SEP achieves the best performance over other pooling strategies, outperforming dRNN using LHS by about 1\% and 1.5\% on KTH-1 and NUS-HGA datasets, respectively. SEP detects the hidden states corresponding to the most intense motions through the energy of DoS. On top of it, mean pooling is applied to suppress the unwanted noise. As a result, SEP can generate both discriminative and reliable representation of states.

\begin{table}
	\centering
		\begin{tabular}{ c | c | c }
		\hline
			 \textbf{Dataset} & \textbf{Method} & \textbf{Accuracy} (\%)\\ 
		\hline
		  \multirow{3}{3.5em}{KTH-1} & LHS & 93.96 \\
			& Mean-Pooling & 94.11 \\
			& Max-Pooling  & 94.29 \\
			& SEP & \textbf{94.78} \\
        \hline
		  \multirow{3}{3em}{NUS-HGA} & LHS & 97.37 \\
			& Mean-Pooling & 97.63 \\
			& Max-Pooling & 97.89 \\
			& SEP & \textbf{98.95} \\
		\hline
		\end{tabular}
	\caption{Performance comparison of SEP against different pooling strategies on the KTH-1 and NUS-HGA datasets. All the results are generated by $2^{nd}$-order dRNN.}
	\label{tab:Pooling}
\end{table}

\begin{figure}
	\centering
		\includegraphics[width=1.0\linewidth]{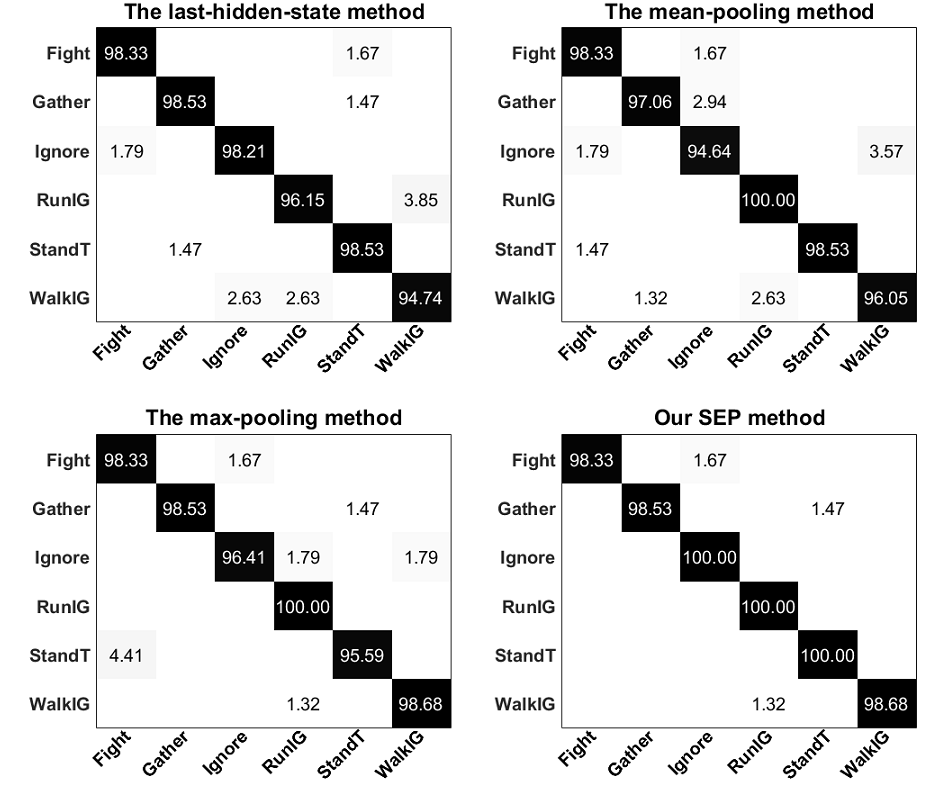}
	\caption{Confusion Matrices of $2^{nd}$-order dRNN models using different pooling methods on the NUS-HGA dataset.}
	\label{fig:cm_pooling}
\end{figure}

To better understand the superiority of State Energy Profile over other pooling methods, we show the confusion matrices of $2^{nd}$-order dRNNs using LHS, mean-pooling, max-pooling and SEP methods on the NUS-HGA dataset in Fig. \ref{fig:cm_pooling}{\color{red}.}. The matrices lead to several interesting discoveries. The LHS method does not achieve 100\% recognition rate in any class. 3.85\% of RunInGroup group behaviors are wrongly classified as WalkInGroup by the LHS method, while all other techniques fully identify RunInGroup behavior correctly. This indicates that the last time-step hidden state cannot sufficiently utilize the DoS, even though DoS better describes the motion pattern. 

The mean-pooling method has large error while dealing with the group behavior Ignore.
This can be explained that mean pooling serves as noise filtering: the information of one person walking by and being ignored by others could be smoothed by mean pooling. The unique motion pattern of the person who is excluded from the group tends to be smoothed by the mean-pooling method, thus, the neural networks get confused by Ignore, WalkInGroup, Gather and Fight behaviors. Our SEP method mitigates the influence of this problem by choosing only hidden states corresponding to intensive and informative motions to generate the final representations. 

Max pooling wrongly classifies StandTalk into Fight by a significant error rate of 4.41\%. Max pooling can select the salient information over hidden states through all time-steps, but is subject to motion noise. When people stand and talk in a group, there might be interactions between them. Max pooling detects such motions and incorrectly identifies them as Fight. StandTalk behavior in the NUS-HGA dataset gives an example of one person raising arm when he is talking. The short and fast behavior could lead to the misclassification. The SEP method performs mean pooling over the candidate hidden states, which largely decreases the chance of incorrect recognition by smoothing the noisy motions.

SEP recognizes the Ignore, RunInGroup and StandTalk behaviors perfectly. This demonstrates that SEP, analyzing the energy of DoS, integrates the virtues of mean pooling and max pooling and minimizes the disadvantages of the two techniques.

\subsection{Run-time Efficiency}

We performed our experiments on a personal computer with an Intel Core i7-6700K CPU, Nvidia GeForce GTX 1080 Ti GPU, and 32GB of RAM. From Table \ref{tab:run_time}{\color{red}.}, we can see that the training convergence takes reasonable amount of time.

Adding up run time for feature extraction and testing per example, dRNN takes 0.414 second averagely to recognize a action on KTH dataset. For the NUS-HGA and BEHAVE datasets, it takes only 0.153 and 0.138 second to identify a group activity, respectively, which fully meets the requirement for real-time applications.

\begin{table}
	\centering
		\begin{tabular}{ c | c | c | c | c }
		\hline
			& \textbf{KTH} & \textbf{MSR}  & \textbf{NUS} (s) & \textbf{VF} (s)\\ 
		\hline
		 Feature extraction & 168 & - & 62 & 82 \\
		 Feature per example & 0.28 & - & 0.13 & 0.13 \\
		 Training convergence & 10340 & 3430 & 2790 & 1090 \\
		 Testing per example & 0.134 & 0.083 & 0.023  & 0.081	\\
		\hline	
		\end{tabular}
	\caption{Run-time efficiency on the datasets.}
	\label{tab:run_time}
\end{table}

\section{Conclusion}
\label{sec:conclusion}
In this paper, we present a new family of differential Recurrent Neural Networks (dRNNs) that extend Long Short-Term Memory (LSTM) structure by modeling the dynamic of states evolving over time. The conventional LSTM is a special form and base model of the proposed dRNNs.
The new structure is better at learning the salient spatio-temporal structure. Its gate units are controlled by the different orders of derivatives of states, making the dRNN model more adequate for the representation of the long short-term dynamics of human activities. Based on the energy analysis of Derivative of States (DoS), we further introduce SEP pooling strategy which can select the most salient hidden states and generate more discriminative representation for video sequences.
Experimental results on human action, group activity, and crowd behavior datasets demonstrate that the dRNN model outperforms the conventional LSTM model. Armed with SEP pooling strategy, dRNN model can further enhance the performance. In the meantime, even in comparison with the other state-of-the-art approaches based on strong assumptions about the motion structure of actions being studies, the proposed general-purpose dRNN model still demonstrates much competitive performance on both single-person and multi-person activity problems.


%





\ifCLASSOPTIONcaptionsoff
  \newpage
\fi



%

\bibliographystyle{IEEEtran}
\bibliography{IEEEabrv,IEEEexample}




%

\begin{IEEEbiography}[{\includegraphics[width=1in,height=1.25in,clip,keepaspectratio]{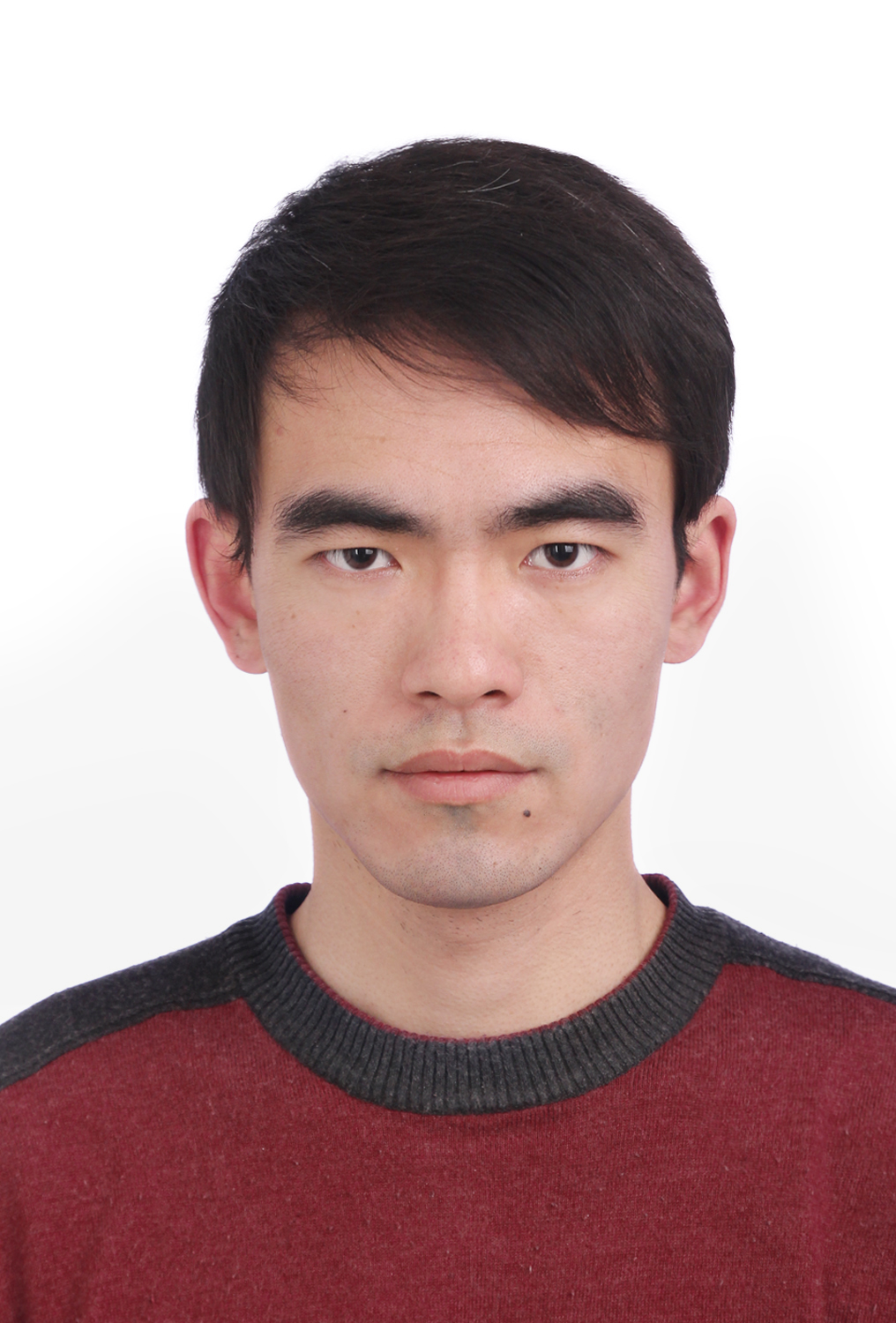}}]
{Naifan Zhuang}
received his B.S. degree in Automatic Control from Northwestern
Polytechnical University in 2008 and M.S. degree in Electrical and Computer
Engineering from University of Florida in 2014. He is currently working towards the Ph.D. degree in the Department of Computer Science, University of Central Florida. 
His research interests include Computer Vision and Deep Learning. His current focus is Deep Differential Recurrent Neural Network for action
recognition and video hashing.
\end{IEEEbiography}

\begin{IEEEbiography}[{\includegraphics[width=1in,height=1.25in,clip,keepaspectratio]{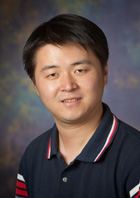}}]
{Guo-Jun Qi} received the PhD degree from the
University of Illinois at Urbana-Champaign, in
December 2013. His research interests include
pattern recognition, machine learning, computer
vision, multimedia, and data mining. He received
twice IBM PhD fellowships, and Microsoft fellowship.
He is the recipient of the Best Paper
Award at the 15th ACM International Conference
on Multimedia, Augsburg, Germany, 2007. He is
currently a faculty member with the Department
of Computer Science at the University of Central
Florida, and has served as program committee member and reviewer for
many academic conferences and journals in the fields of pattern recognition,
machine learning, data mining, computer vision, and multimedia.
\end{IEEEbiography}

\begin{IEEEbiography}[{\includegraphics[width=1in,height=1.25in,clip,keepaspectratio]{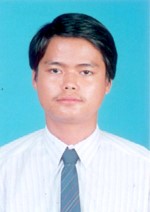}}]
{The Duc Kieu}
received the B.S. degree in Mathematics from the University of Pedagogy, Vietnam, in 1995, the B.S. degree in Information Technology from the University of Natural Sciences, Vietnam, in 1999, the M.S. degree in Computer Science from Latrobe University, Australia, in 2005, and the Ph.D. degree in Computer Science from Feng Chia University, Taiwan, in 2009. Since 2010, he has been with the Department of Computing and Information Technology, Faculty of Science and Technology, The University of the West Indies, St. Augustine, Trinidad and Tobago, where he is currently a tenured Senior Lecturer. His research interests include information hiding, data compression, image processing, and machine learning.
\end{IEEEbiography}

\newpage
\begin{IEEEbiography}[{\includegraphics[width=1in,height=1.25in,clip,keepaspectratio]{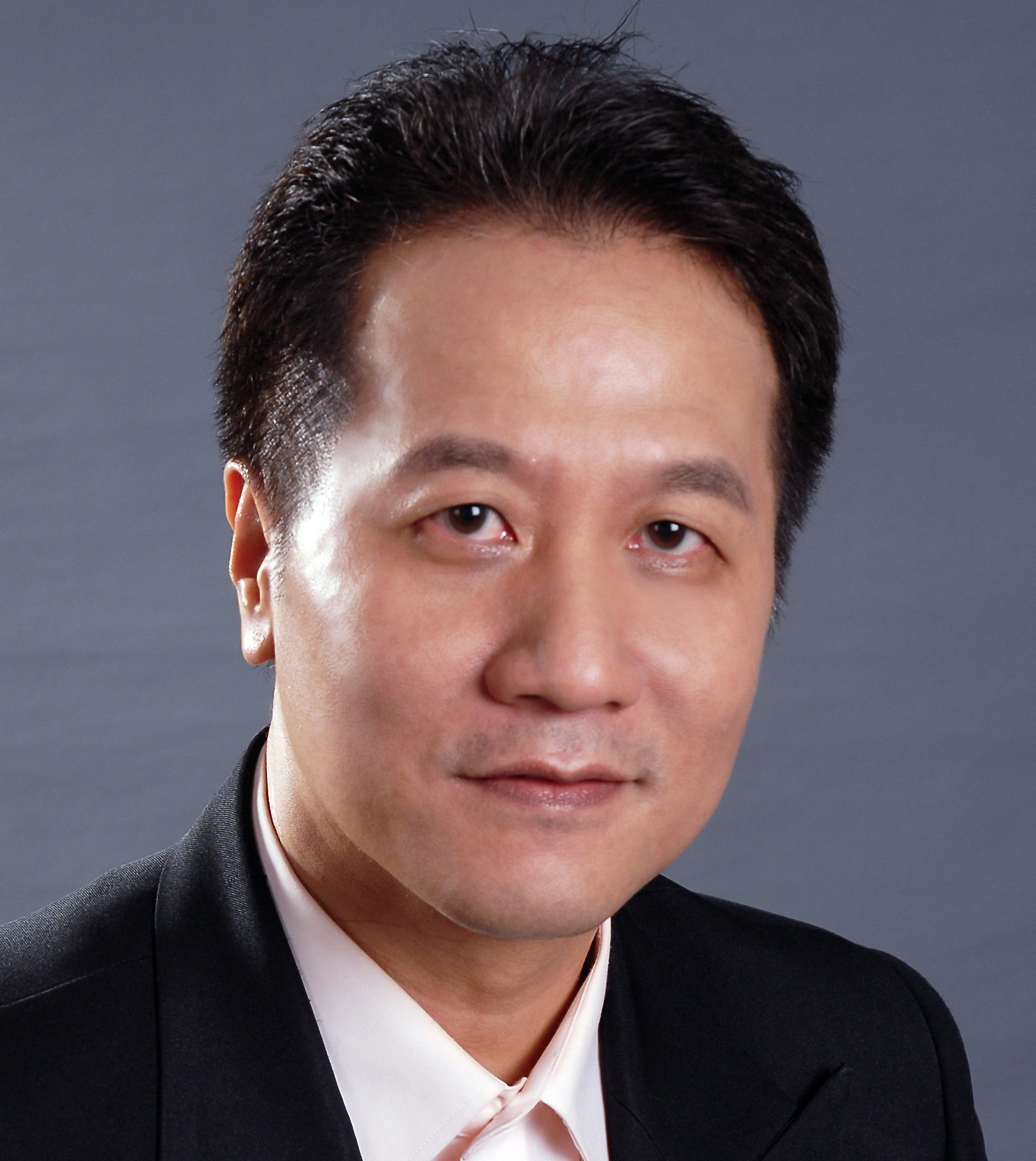}}]
{Kien A. Hua}
is a Pegasus Professor and Director of the Data Systems Lab at the University of Central Florida.  He was the Associate Dean for Research of the College of Engineering and Computer Science at UCF.  Prior to joining the university, he was a Lead Architect at IBM Mid-Hudson Laboratory, where he led a team of senior engineers to develop a highly parallel computer system, the precursor to the highly successful commercial parallel computer known as SP2.  Dr. Hua received his B.S. in Computer Science, and M.S. and Ph.D. in Electrical Engineering, all from the University of Illinois at Urbana-Champaign, USA.  His diverse expertise includes multimedia computing, machine learning, Internet of Things, network and wireless communications, and mobile computing.  He has published widely with 13 papers recognized as best/top papers at conferences and a journal.  Dr. Hua has served as a Conference Chair, an Associate Chair, and a Technical Program Committee Member of numerous international conferences, and on the editorial boards of several professional journals.  Dr. Hua is a Fellow of IEEE.
\end{IEEEbiography}








\end{document}